# COVID19 UK Bounce Back Loan Scheme

## An Efficacy Assessment of Pandemic Driven Business Support Initiative

**INM433 Visual Analytics**

**Arunav Das**

arunav.das@city.ac.uk

City University, London, UK

**Abstract**— *Bounce Back Loan is amongst a number of UK business financial support schemes launched by UK Government in 2020 amidst pandemic lockdown. Through these schemes, struggling businesses are provided financial support to weather economic slowdown from pandemic lockdown. £43.5bn loan value has been provided as of 17$^{th}$ Dec2020. However, with no major checks for granting these loans and looming prospect of loan losses from write-offs from failed businesses and fraud, this paper theorizes prospect of applying spatiotemporal modelling technique to explore if geospatial patterns and temporal analysis could aid design of loan grant criteria for schemes. Application of Clustering and Visual Analytics framework to business demographics, survival rate and Sector concentration shows Inner and Outer London spatial patterns which historic business failures and reversal of the patterns under COVID-19 implying sector influence on spatial clusters. Combination of unsupervised clustering technique with multinomial logistic regression modelling on research datasets complimented by additional datasets on other support schemes, business structure and financial crime, is recommended for modelling business vulnerability to certain types of financial market or economic condition. The limitations of clustering technique for high dimensional is discussed along with relevance of an applicable model for continuing the research through next steps.*

---

## 1 Problem Statement

Could application of Visual Analytics have led to a different economic outcome from UK's Bounce Back Loan Scheme (BBLS)?

BBLS was launched in April 2020 [1] for supporting Small, Medium sized Enterprises (SMEs) impacted by Coronavirus pandemic lockdown. Until scheme expiry [2], UK businesses from any location/sector could apply for loan of £2,000-£50,000 (maximum 25% business' turnover) from Accredited Lenders [3]. £43.5 billion [4] is estimated to have already been granted from which £15-£26 billion loan losses [5] are estimated by National Audit Office due to failure of businesses to pay back or fraudulent loan applications.

Aside from business owner self-certification, traditional credit checks were bypassed for processing BBLS applications. Whilst bypassing traditional checks could have been necessary to expedite support for businesses facing challenges, potential losses from future write-offs, frauds might have been mitigated through an analytical decision making involving historical business failure rates, sector concentration by reviewing pattens for

- What % of businesses fail every year i.e., would some businesses fail even without COVID19?
- Does business death rate change by UK Region, Sector, Time?
- What Regions, Sectors are more/less impacted by lockdown and require more/less support from BBLS

Information needed for this research is available under - ONS[1] historical spatiotemporal datasets [12] with UK business demographics (Birth, Death, Active -Location, Sector); spatial dataset [12,13] with Sectorial information by location; BBB[2] spatiotemporal dataset for BBLS [14]; ONS[1] business sentiment [15] survey.

Visual Analytics applied to Baseline and Comparative datasets is expected to provide insights into the efficacy of BBLS

## 2 State of the Art

Whilst research on role of visual analytics for government-backed financing schemes nor business lending assessments by banks is available, there is evidence for application of visual analytics on spatiotemporal datasets of mortgages, network-guaranteed loans, government bonds to aid lending and risk management decisioning.

Heinen et al [6] have modelled mortgage defaults over geographic distances in Los Angeles using spatiotemporal dataset from 2000-2011 for demonstrating geospatial dependence between loan defaults and zip codes. Their approach involves grouping mortgages into risk categories of prepayment vs default by applying multinomial logistic regression model followed by Copulas method linking mortgage defaults to zip codes using Matern correlation function. Timeseries analysis helped decisioning for splitting dataset into pre-/post-2005 for considering higher default rates post 2005 due from introduction of new mortgage products. Their model shows high dependency of defaults within 40 km geographic range due to shared socioeconomic factors.

---

1. *Office for National Statistics, UK*
2. *British Business Bank*

Niu et al. [7] proposed new way of managing financial systemic risks from business borrowing needs through a network-guarantee framework and segregation of high-default groups. Their approach involves multi-faceted risk visualization interface linking default risk prediction made through gradient boosting tree over the artificial spatial features of the business network using network centrality measures across 400+ nodes and 103 enterprises.

Szulc et al. [8] et al used spatial correlation to assess impact of credit rating changes on government bonds on the yield on government bonds of neighboring countries and found a statistically significant negative correlation using dynamic spatial panel models on grouped timeseries and cross-sectional data of ten-year government bonds and economic factors (e.g., inflation, volatility, rating increase/decrease) across 40 countries for 2008-2017 timeseries. Traditionally, the yield on government bonds have only been evaluated through credit ratings. Their research shows a geospatial correlation of government bond yields

More generally, Adrienko(s) [9] discuss issues with standalone computational methods and interactive visualization for modelling real-world spatiotemporal datasets. They propose a new framework for modelling such datasets through a combination of clustering multiple spatial timeseries and interactive grouping of geographic objects and spaces. The proposed framework includes use of animated cartographic displays, time graphs, interactive tools for cluster analysis and statistical methods for modelling timeseries data applied on transformed, grouped and filtered spatiotemporal datasets. Yao [10] addresses similar issues by proposing an analysis framework that involves slicing spatiotemporal datasets into Spatialization and Temporalization categories for segmentation, dependency analysis, outlier detection and trend discovery.

Leveraging the specific risk management frameworks and generic visualization techniques for spatiotemporal datasets from literature review, this paper applies Visual Analytics on Spatialization and Temporalization framework to analyze timeseries of business demography, Sector concentration alongside spatial distances/transformation of parliamentary constituencies to compare against the spatial distribution and size of BBLS in UK for evaluation of applied Visual Analytics to propose amendment to government BBLS Scheme policy/bank's lending criteria with potential benefits linked to lower write-off probability from bad loans. Similar Policy review aspect of application of Visual Analytics on spatiotemporal data has been presented by Wood et al [2010] using socio-economic datasets from Leicestershire.

## 3 PROPERTIES OF THE DATA

### 3.1 Data Collection & Sources

#### 3.1.1 Business & Sectoral Demographics – UK & London

Information published by ONS [12] for annualized number of Enterprise Births, Deaths, Survivals from 2014-2019 has been used as Baseline Dataset. Geographically, this information is split by UK District, Counties, Regions and by Industries. This provides insight on sectoral and regional composition and concentration of enterprises listed with – annual number of new, failed, active businesses.
London Borough's business demographic dataset published by ONS [13] has been used for detailed analysis of Greater London region over an extended period of 2004-2018

#### 3.1.2 BBLS Loan Applications

Information about UK Government's pandemic support initiatives for businesses is available from Commons Library [14]. This report contains point in time, 17th December 2020, information about total number of loan applications and value of loans (£) granted since its inception under BBLS, split-by Parliamentary Constituencies. This information has been used as Comparative Dataset for evaluating efficacy of BBLS through by comparing pre-pandemic regional and sectoral business demise with post-pandemic regional/sectoral BBLS Loan distribution to identify spatial distribution of regions/sectors with historic high/low business deaths vs BBLS loan distribution

#### 3.1.3 Business Sentiment Analysis

COVID19 Business Impact survey results published fortnightly by ONS [15] used for comparing BBLS sectoral loan distribution with sectoral sentiments. Dataset contains % businesses with trading paused by sector

## 3.2 Data Characteristics & Quality

### 3.2.1 Temporal Data Analysis

UK and London Business demographics dataset have a timeseries component that shows annual variation of number of new, failed and active businesses by geographic location as well as industries across 5- and 14-years horizon respectively. Business sentiment analysis has a timeseries component that captures business feedback since 9th March 2020 with a fortnightly frequency

Violin and Line plots [Fig1] of UK business demographics timeseries shows - 1. unimodal distribution of business survival/death/birth with higher densities at lower values with long tails representing a lower density of high number of business demise/births and 2. upward trend in new businesses till 2016 followed by an upward trend in business demise from 2017

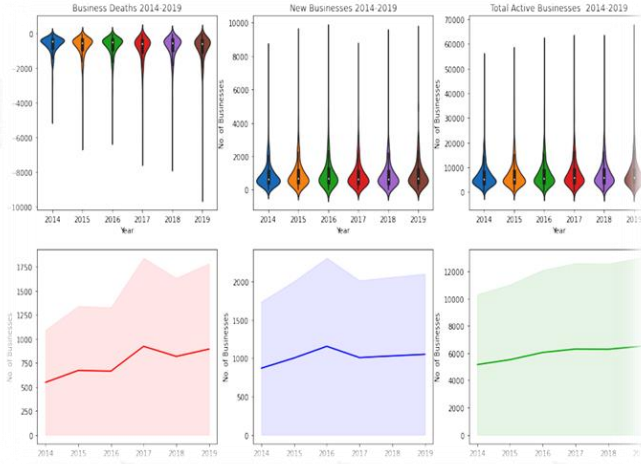

**Fig1**

### 3.2.2 Spatial Data Analysis

UK Business demographics dataset provides information for grouped by 12 regions of UK.
Geographic locations for England include name and geocodes for 48 ceremonial, 6 metropolitan and 77 non-metropolitan counties, 314 districts, 224 non-and-metropolitan districts, 33 London Boroughs, 56 unitary authorities [24]. Name and geocode information for Scotland, Wales and Northern Ireland contain 32, 22 and 11 areas/districts respectively. London Business Demographic data contains information related to 33 Boroughs. BBLS data (numbers and £value) is reported through 650 Parliamentary Constituencies

The resulting artificial space for Sector classification contains 82 sectors with 252 sub-sectors grouped into 17 SIC[5] for linking number of business births/deaths/active to respective geographic locations

Cartographic encoding of raw data [Fig2] does not immediately help to spot spatial correlation between historic trends of business deaths with loans granted through BBLS. Also, there are exceptions related to unmapped locations implying need for cleaning and mapping geocode locations as they have changed over a period of time.

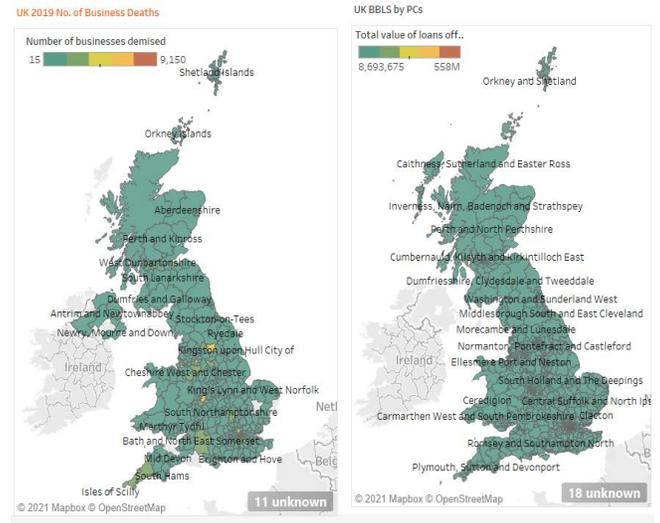

**Fig2**

## 4 ANALYSIS

### 4.1 Approach

Diagram1/Appendix1 represents four-stage research with iterative human-machine interaction as outlined below

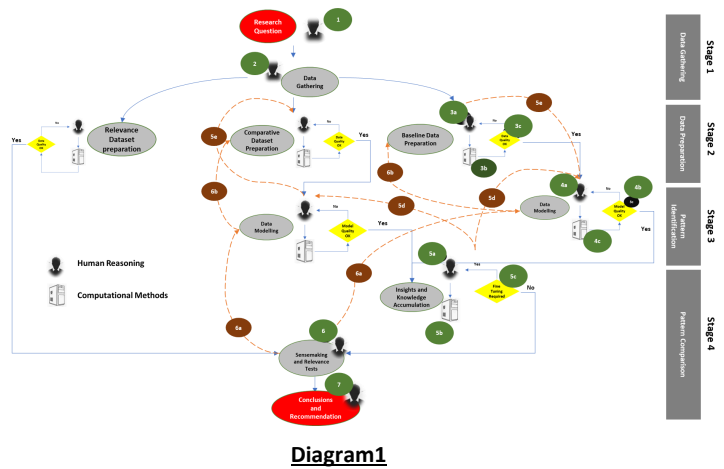

**Diagram1**

*Stage1: Data Gathering* [steps 1, 2, Diagram1]

Identification of relevant datasets was carried out through Google searches (keywords – UK Business Demography,

---

3. [Financial support for businesses during coronavirus (COVID-19) - GOV.UK (www.gov.uk)](#)
4. Coronavirus Business Impact Loan, 5. Standard Industry Classification

BBLS, business sentiments post-COVID19). Purpose of this step was to assess availability of datasets for continuing research. Search results were categorized into Baseline, Comparative and Relevance Testing datasets. Motivation for dataset segregation was to conduct spatiotemporal analysis on Baseline dataset for comparing spatial features with Comparative dataset and locating discernible patterns between BBLS applications vs historic spatial patterns of annual business deaths.

Human-Machine interface consisted of an iterative process of rephrasing the search criteria by applying human reasoning to keywords depending on relevance of datasets found through previous google searches.

*Stage2: Data Preparation* [steps 3a,3b,3c-Digram1]

Business demographics data available separately for 2014, 2015-2018 and 2019 was merged to create consolidated database. Exceptions identified during this step was manually corrected. Regional and country information was manually deleted from consolidated dataset, only county/district information was retained. Quality of merged database was checked using Violin and Timeseries plots. Outliers highlighted by Violin-plot were manually reviewed to assess if they should be retained or removed. Spatial data quality check was performed by plotting geocodes. 10 Districts were found as unrecognized. These geocodes were manually remapped.

Comparative dataset was created by extracting BBLS dataset from year-to-date Total Loans data. Violin plot was used to review data distribution and outliers. Spatial data quality check highlighted 18 unmapped Parliamentary Constituencies.

Consolidated database created for Relevance dataset by merging fortnightly business sentiment analysis surveys to create a timeseries

Human-Machine interface included use of iterative computational methods for (re)creating merged datasets and statistical plots. Human reasoning was used to resolve unmapped locations and reviewing patterns, null values and outliers to resolve underlying issues.

*Stage3: Pattern Detection* [steps 4a,4b,4c-Diagram1]

K-Means clustering method was applied to the Baseline data for 2019 business demographics to identify clusters with low, medium and high rates of business failures across various districts and industries. K-Means clustering was also applied to the Comparative dataset to identify low, medium and high BBLS loan amount per Parliamentary Constituencies. These clusters were used for comparing locations with different rates historic business failure vs locations granted with BBLS loans.

Different clustering techniques were compared for efficacy.

Human-Machine interface included human reasoning to design the clustering model and computational methods for executing the clustering algorithms. Human reasoning was then used to fine tune the model for better quality clusters.

*Stage4: Pattern Comparison* [steps 5,6,7-Diagram1]

Insight from clusters was checked against Business Sentiments for assessing relevance of visual analytics for recommendation of BBLS eligibility criteria

Based on the clustering analysis, specific region was chosen to conduct a deep-dive analysis of the factors impacting BBLS applications [5d,5e-Diagram1].

Human-Machine interface included human reasoning for causal analysis of observed clusters. Computational techniques were applied to aid this step by plotting timeseries heatmap of business deaths for extrapolating non-Covid19 business deaths in 2020 and through analysis of business sentiments survey [6a,6b-Diagram1]

**4.2 Process**

*4.2.1 Data Preparation*

Outliers analysis shows that very high business death rates and very high number of BBLS applications are related to a few counties/regions like London, Manchester and Birmingham. As all the regions are important to address the research question, these outliers have been retained for pattern recognition analysis.

BBLS data with "Constituency unspecified" was removed to address null value for geocoding.

New fields derived from Baseline dataset for

i. Business Survival Index= business death/business birth

ii. Death Ratio Index = business death/total active business for temporal trend analysis of baseline data

iii. Average business deaths/births/active

New field derived in Comparative dataset for

iv. Average BBLS Loan = Total Loan Value/No. of Applications

As part of business sentiment analysis, 17 different reports from April-November 2020 were consolidated to create a sectoral timeseries with information related to sectors that paused trading during lockdown

*4.2.2    Spatial Analysis & Pattern Identification – UK*

Clustering - an exploratory unsupervised machine learning technique, involves mining datasets for grouping similar/ dissimilar elements [16]. Clustering of spatiotemporal datasets aid pattern recognition by enabling spatial feature comparison of various clusters as well as study of specific cluster for analyzing impact from temporal dimension [17,18,19]. Hence clustering technique was applied to compare patterns between Baseline and Comparative Datasets.

K-means partition-based clustering method was chosen for this research over Hierarchical (HC) and Density Based Clustering (DBS) because of importance for including outliers. K-Means method includes all datapoints for clustering whereas DBS could potentially group outliers as noise, distorting pattern analysis. Hierarchical clustering similarly only considers grouping close datapoints without due consideration for features. Though it could be used for embedded cluster pattern detection in future as extension of this research.

As K-Means method is not well suited for high dimensional dataset [20], Principal Component Analysis (PCA) [25] was used to reduce business demographics dimensions from 18 features to 2 [Fig3a] with 96% Explained Variance and Sector classification from 16 to 2 features. Elbow method [21] was used on reduced dimensional dataset to identify optimum cluster size as 3 [Fig3b]

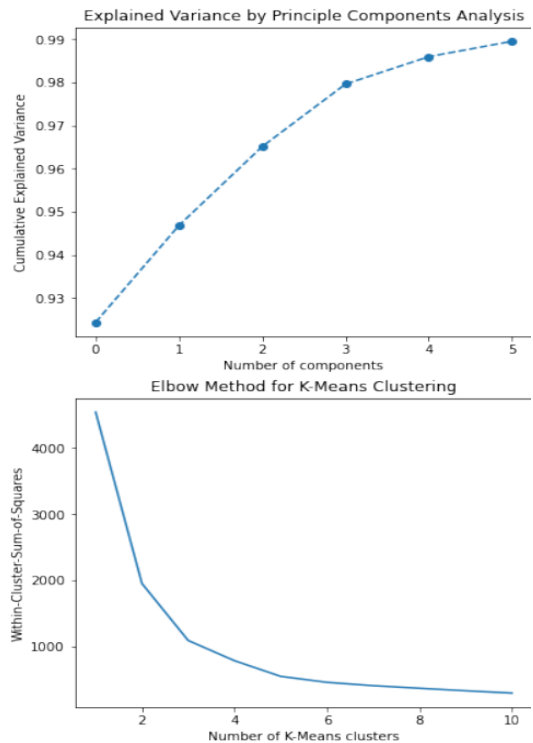

**Fig3a,b**

K-Means clustering was run on - a. Business demographics information, b. Sector classification, c. BBLS information

*4.2.3    Cluster Evaluation - UK*

Combination of higher Silhouette and lower David Boulding scores were used for cluster quality assessment and clustering method selection. Overall, three clusters were chosen for Business Demographics and Sector Classification datasets due to superior comparative scores. Three clusters scenario was chosen for BBLS for equitable spatial comparisons across all three datasets even though 2-cluster scores were higher for BBLS [Table1]

Summary of the K-Means clustering iterations

| a. Business Demography | PCA=2, Cluster = 2 | PCA=2, Cluster = 3 | PCA=2, Cluster = 4 | PCA=3, Cluster = 3 |
|---|---|---|---|---|
| Silhouette Score | 0.73 | 0.703 | 0.6 | 0.69 |
| David Bouldin Score | 0.58 | 0.55 | 0.58 | 0.58 |
| b. Sector Classification | PCA=1, Cluster = 2 | PCA=1, Cluster = 3 | PCA=2, Cluster = 2 | PCA=2, Cluster = 3 |
| Silhouette Score | 0.71 | 0.68 | 0.66 | 0.61 |
| David Bouldin Score | 0.55 | 0.48 | 0.74 | 0.72 |

3. [Financial support for businesses during coronavirus (COVID-19) - GOV.UK (www.gov.uk)](www.gov.uk)
4. Coronavirus Business Impact Loan, 5. Standard Industry Classification

|  | Cluster1 | Cluster2 | Cluster3 |
|---|---|---|---|
| Business Demography<br><br>Total Districts and Counties = 382 | No. of members 316<br><br>5-year Avg. failed business 535, Low | No. of members 58<br><br>5-year Avg. failed business 1690 Medium | No. of members 8<br><br>5-year Avg. failed business 4134, High |
| Sector Classification<br><br>Total Districts and Counties = 389 | No. of members 324<br><br>Low Avg. No of Industries<br><br>Top 3 Industries Professional, Business Admin., Construction | No. of members 61<br><br>Medium Avg. No. of Industries<br><br>Top 3 Industries Professional, Construction, Information-Communication | No. of members 4<br><br>High Avg. No. of Industries<br><br>Top 3 Industries Professional, Business Admin., Information-Communication |
| BBLS distribution<br><br>Total Parliamentary Councils = 644 | No. of members 334<br><br>Avg. No. of applications 1386, Low | No. of members 251<br><br>Avg. No. of applications 2139, Medium | No. of members 59<br><br>Avg. No. of applications 3678, High |

| c. BBLS (PCA Not Required) | Cluster = 2 | Cluster = 3 | Cluster = 4 | Cluster = 5 |
|---|---|---|---|---|
| Silhouette Score | 0.64 | 0.56 | 0.53 | 0.52 |
| David Bouldin Score | 0.57 | 0.56 | 0.54 | 0.53 |

**Table1**

K-Means clustering results are compared with DBS and Hierarchical Clustering results [Appendix2]. Based on the Silhouette and David Bouldin Scores, K-Means clustering returns best results. Hence, K-Means was chosen as the clustering method for this research. Dataset features from resulting clusters were visually reviewed for understanding attributes of cluster composition [Table2]. Cluster 3 was found to represent business hotspots with historic high rate of failed, new businesses, Sector concentration as well as high BBLS loan applications.

Geocode-mapping of clusters [Fig4a/b/c] shows high impact UK Regions are London with 5, North West with 3 and South East 2 Parliamentary Constituencies with higher business death rates and BBLS applications.

Whilst top three Regions from Cluster3 could all be studied individually, due to maximum number of impacted Parliamentary Constituencies (Fig4d), London Region was chosen for further analysis to study discernible correlation between patterns of historic business death rates, Sector concentration and BBLS applications through identification of geospatial clusters and impact from temporal trends

**Table2**

Geocode mapping errors [Appendix7] for mapping locations during this step were manually resolved through look-ups of missing geolocations. The errors were found to be related to changes to the name of certain districts on the source datasets (e.g., Northern Ireland) since 2015 onwards

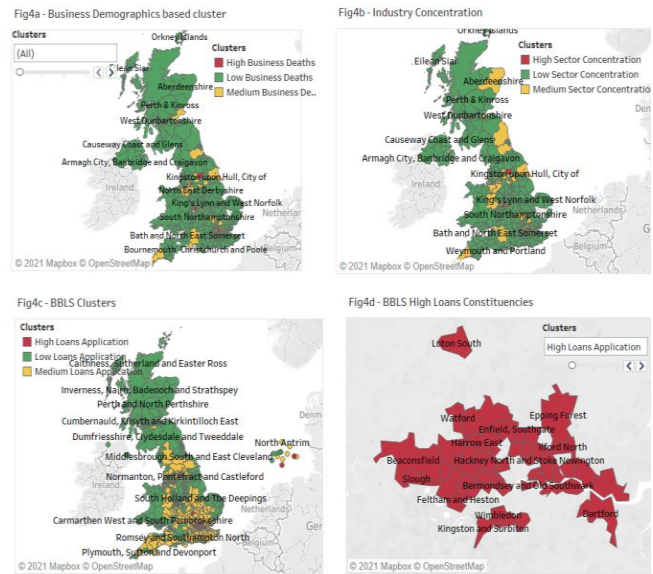

**Fig4a-b-c-d**

4.2.4 Pattern Identification & Cluster Evaluation – London Region

London data was extracted from Baseline and Comparative datasets. Information for Districts, 73 Parliamentary Constituencies (PC) from the underlying datasets were consolidated for mapping to 33 London Boroughs to create comparable clusters across for pattern identification across all datasets. Pattern extraction via Regional clustering was basis for this consolidation

K-Means clustering was rerun. Quality checks for optimum number of clusters [Appendix3] were performed by changing the number of clusters and measuring combination of higher Silhouette with lower David Bouldin score. Three clusters were found as optimum for all datasets. Even though clusters returned one member for two datasets, these clusters were retained due to impact from its features. Based on visual feature and cluster membership analysis for each dataset a Cluster Summary [Table3] was created with distinction for High-Medium-Low impact of business deaths, Sector concentration and BBLS applications.

|  | Cluster1 | Cluster2 | Cluster3 |
|---|---|---|---|
| **Business Demography** <br> Total Boroughs = 33 | No. of members <br> 9 <br><br> 5-year Avg. failed business <br> 2946, Medium | No. of members <br> 1 <br><br> 5-year Avg. failed business <br> 6397, High | No. of members <br> 23 <br><br> 5-year Avg. failed business <br> 1582, Low |
| **Sector Classification** <br> Total Boroughs =33 | No. of members <br> 24 <br><br> Low Avg. No. of Industries <br><br> Top 3 Industries <br> Professional, Information-Communication, Business Administration | No. of members <br> 1 <br><br> High Avg. No. of Industries <br><br> Top 3 Industries <br> Professional, Information-Communication, Property | No. of members <br> 8 <br><br> Medium Avg. No. of Industries <br><br> Top 3 Industries <br> Professional, Information-Communication, Retail |
| **BBLS distribution** <br> Total Boroughs =33 | No. of members <br> 13 <br><br> Avg. No. of applications & Loan amount <br> 8547, £32.8k, Medium | No. of members <br> 14 <br><br> Avg. No. of applications & Loan amount <br> 4957, £32.4k, Low | No. of members <br> 6 <br><br> Avg. No. of applications <br> 13189, £34.9k, High |

**Table3**

### 4.2.5 Pattern Comparison – London Region

To compare spatial patterns of historic business death rate and Sector concentration with BBLS Loan applications across all 33 London Boroughs, maps were plotted as per the three clusters [Appendix4] with every Borough identified as one of High-Medium-Low impacts separately for each dataset type.

Cluster filters were used to separate the Boroughs with highest demand (average 13k applications, marked in "red") for BBLS [Fig5a], highest rate of business failures [Fig5b, marked in "red"] and highest activities as per Sector concentration [Fig5c, marked in "red"].

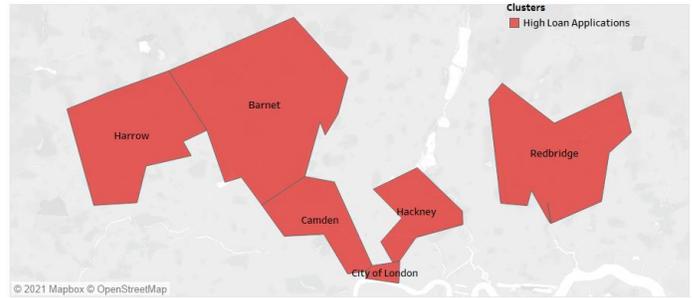
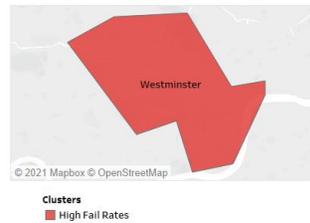
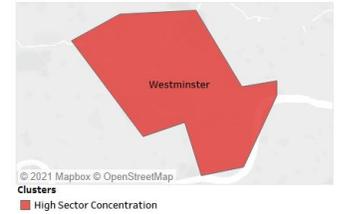

**Fig5a-b-c**

Similar cluster filters were used to separate out medium sized (avg. 8k, marked in "amber") applications [Fig6a-b-c] for spatial comparison of Boroughs with higher/lower historic business failures vs. BBLS applications

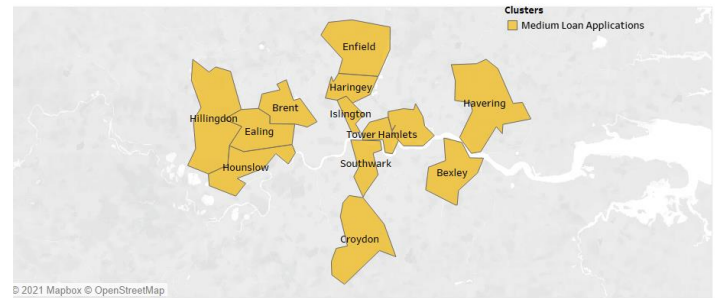
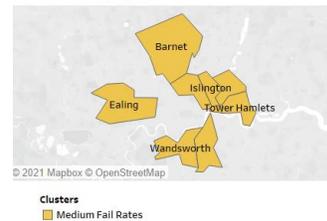
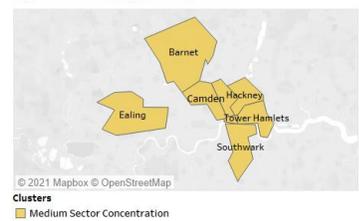

**Fig6a-b-c**

### 4.2.6 Temporal Analysis of Business Demography – London Region

To support spatial analysis through temporal dimension, 2004-2019 timeseries of Net (=new – failed) Business Growth was plotted (Fig7a) for the High-Impact Boroughs to review underlying annual business survival rates and identify any discernible positive/negative trends that might have anyway caused problems to businesses in 2020 even without the pandemic. This highlighted 2009 and 2017 as years when

---

3. Financial support for businesses during coronavirus (COVID-19) - GOV.UK (www.gov.uk)
4. Coronavirus Business Impact Loan, 5. Standard Industry Classification

London Boroughs saw higher rate of business deaths. 2017 was chosen due to its temporal proximity to 2020 to plot TreeMap for visualizing correlations between the size of 2017 business death per Borough in 2017, anomaly evaluation comparing it with the timeseries growth rate in 2016 and subsequent recovery by 2019. This aids in reviewing the temporal variation of business death rates by Boroughs and its correlation with BBLS applications.

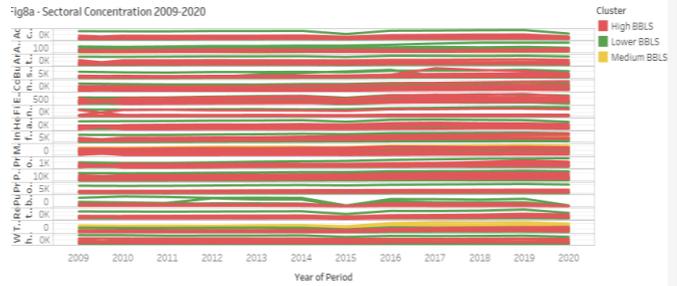

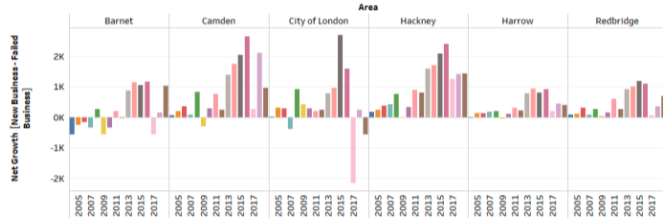

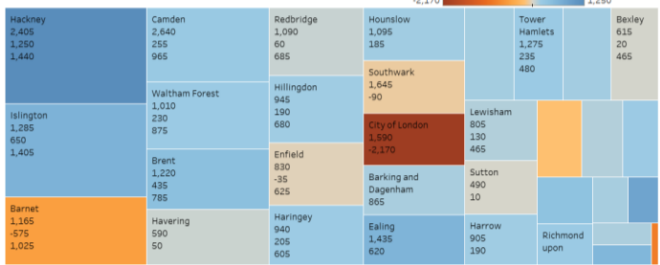

**Fig7a-b**

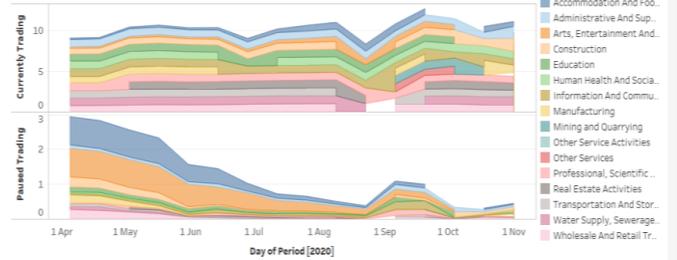

**Fig8a-b**

Insights from Sectoral and Sentiment timeseries was supplemented with London sectoral concentration analysis [Fig9/Appendix6] through 17-Sector histograms, color encoded by cluster labels, and marked as 'High(red)-Medium(amber)-Low(green)' to identify Boroughs with high concentration and high vulnerability from COVID19 business impact.

Similar timeseries was plotted for Medium-Impact Boroughs [Appendix 5]

4.2.7 Temporal Analysis of London Sectoral Concentration & post lockdown Business Sentiments

2009-2020 Timeseries analysis [Fig8a] of Sectoral concentration by Boroughs and encoded by BBLS spatial cluster colors exhibit decline in specific sector concentrations. E.g., Accommodation, Public Concentration, Retail, Wholesale concentration declined in Westminster in 2015 and growing again in 2019. Similarly, sectoral concentration appears to grow in other Boroughs. E.g. Camden with growth in Accommodation, Education and Production. Such a timeseries is useful to understand the temporal variation of Sector concentration for researching its correlation with BBLS applications.

Sectoral Timeseries [Fig8b/Appendix8] of Trading Status for Businesses responding to COVID19 Sentiment Survey shows Accommodation, Arts/Entertainment, Construction, Manufacturing, Retail sectors with highest rate of pausing trading implying sectors that required support to survive.

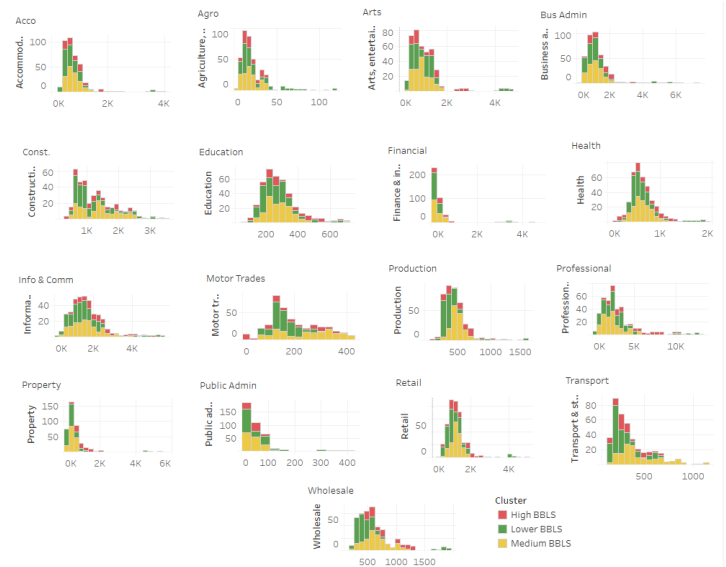

**Fig9**

Features of geospatial clusters was analyzed through these two timeseries outputs to understand correlation between historic death rates, sector concentration with spatial BBLS clusters.

## 4.3 Results

Geospatial cluster analysis shows businesses in Outer London (OL) have been disproportionately impacted by COVID19 lockdown. Until last year 9 OL Boroughs, highlighted in Table 4a, had comparatively lower business death rates. However, BBLS applications from these Boroughs have been higher. Barnet is an exception in Outer London as its BBLS applications appear to be proportional to historic timeseries trends of business deaths [Fig8a].

| Outer London | 2020 BBLS applications | 2014-2019 Business Death Rates | 2019 Sector Concentration |
|---|---|---|---|
| Barnet | | | |
| **Harrow** | | | |
| **Redbridge** | | | |
| **Bexley** | | | |
| **Brent** | | | |
| **Croydon** | | | |
| Ealing | | | |
| **Enfield** | | | |
| **Havering** | | | |
| **Hillingdon** | | | |
| **Hounslow** | | | |

**Table 4a**

For Inner London, 2 Boroughs [Table4b] appear to be outliers with higher BBLS applications. On the contrary, Westminster has lower proportion of BBLS applications compared to historic trends. Like Barnet, timeseries trends supports the higher proportion of BBLS applications for City of London

| Inner London | 2020 BBLS applications | 2014-2019 Business Death Rates | 2019 Sector Concentration |
|---|---|---|---|
| Camden | | | |
| City of London | | | |
| Hackney | | | |
| **Haringey** | | | |
| Islington* | | | |
| **Newham** | | | |
| Southwark | | | |
| Tower Hamlets | | | |
| Westminster | | | |

**Table4b**

Based on Temporal analysis [Fig10] of these 11 highlighted Outer/Inner Boroughs, disproportionate BBLS application appears to be due to a combination of sector concentration and historic death rates. For e.g., Wholesale, Retail, Entertainment, Construction, Accommodation & Food Services and Recreation & Health

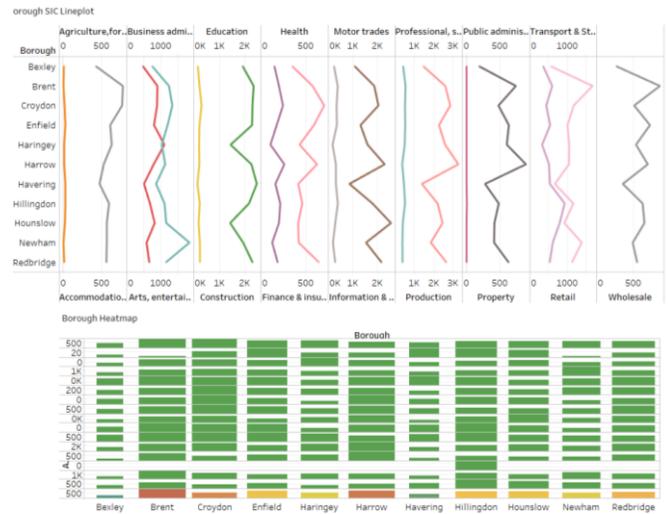

**Fig10**

## 5 CRITICAL REFLECTION

Research motivation was to locate geospatial patterns to aid decision-making process for government backed financing schemes. Such patterns would help to model lending decisions by segregating good businesses from bad business models and help building a decision support system enabling sustainable financing for economic growth model.

Initial application of clustering technique to UK-wide business demographics and sector concentration helped identifying three UK regions with high historic business failure rate. London Region was chosen from this step for comparing business death rates geo-spatiotemporal patterns with BBLS applications.

Visual Analytics applied to London Boroughs clearly identified geospatial patterns contrasting Inner from Outer Region Boroughs for business death rates. However, when measured through BBLS grants, Visual Analytics identified Outer Boroughs with comparatively greater business impact from COVID19 than Inner Boroughs. This spatial discrepancy between pre- and post-pandemic business support need indicates standalone clustering technique needs to be supplemented with other parameters for defining future BBLS eligibility criteria. Histogram plot of Sector information [Fig8] encoded by cluster labels might mitigate help by identifying concentration patterns of pandemic vulnerable sectors in UK. However, the same sector (e.g., Retail) have been impacted differently during COVID19 depending on operating channel (e-Commerce vs Stores) and should be an additional consideration.

Business Demographics Temporal analysis identifies 2009, 2017 with higher annual business deaths. Whilst 2009 is due to aftereffects from 2008 Financial Crisis, 2017 should be reviewed further (e.g., 2016 Brexit-referendum impact). Sectoral Temporal Analysis identifies 2015 with dip for specific sectors (e.g., Accommodation, Retail). This exhibits the importance of temporal insights alongside geospatial patterns. However, availability of only annual temporal data doesn't help in analysis of seasonality.

---

3. Financial support for businesses during coronavirus (COVID-19) - GOV.UK (www.gov.uk)
4. Coronavirus Business Impact Loan, 5. Standard Industry Classification

Iterative Visual Analytics methodology with geospatial and temporal insights related to businesses has proved valuable for identifying patterns through clustering thereby narrowing focus of research from pattern recognition at National level to specific Boroughs in London. Understanding of cluster features would help in building future decision models.

Robustness of identified spatial patterns could be improved by adding datasets from Government COVID19 Financial support[3] *(e.g., Furlough, CBIL[4]),* Business Models (e.g., turnover, distribution channels, workforce size), changes in Consumer spending habits and Financial Crime (e.g., Fraud, Money Laundering) datasets. However, adding more dimensions would make clustering algorithm less effective. Combination of Support Vector Machine with Guassian Process for classification alongside Kohonen Self-Organising MAP as visualization tool could be used for mitigating the 'Curse of Dimensionality' [23]

A Multinomial and Spline Regression model [22] could be built from analysis of cluster features for government grants decision making. Such models are used for Credit Risk modelling in Retail Bank. The model could be trained on London datasets and tested on the Manchester and Birmingham spatiotemporal datasets that represents similar level of business activities and BBLS applications. As the BBLS scheme date has been extended till Jan2021, also depending on pace of economic recovery post pandemic, this proposed model could help in designing future financing schemes for supporting businesses based on spatial patterns, temporal trends, sectoral concentrations, business structures & employees, consumer sentiments and financial crime datasets.

---

3. Financial support for businesses during coronavirus (COVID-19) - GOV.UK (www.gov.uk)
4. Coronavirus Business Impact Loan, 5. Standard Industry Classification

## APPENDIX 1: APPROACH STAGES WITH STEPS

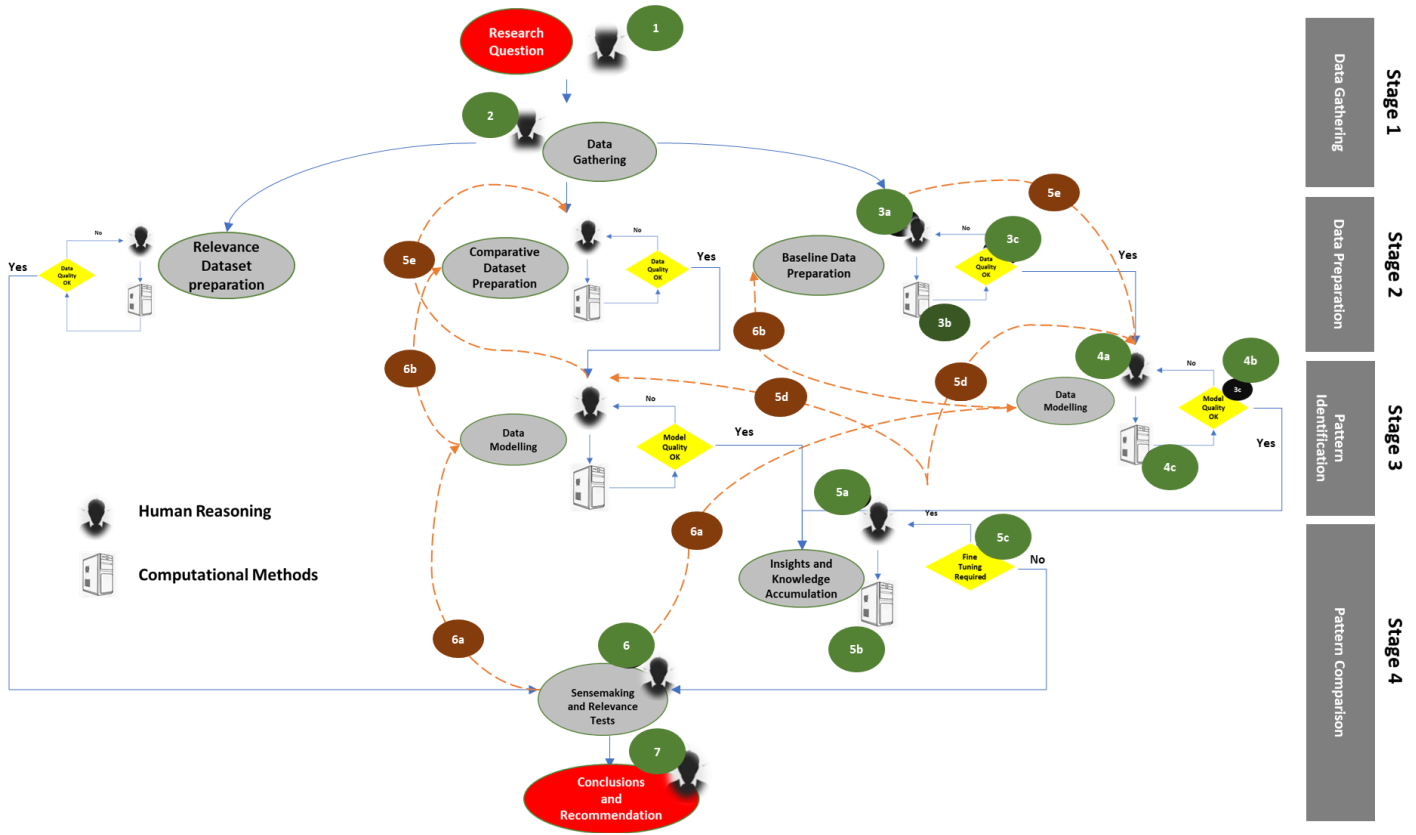

APPENDIX2 : CLUSTER QUALITY ASSESSMENT SCORES
FOR ALTERNATIVE METHODS – HIERARCHICAL AND
DENSITY BASED CLUSTERING TECHNIQUES

| Hierarchical Clustering | | Density Based Clustering | |
|---|---|---|---|
| **a. Business Demographics** | **PCA=2, Cluster= 2** | a. Business Demographics | PCA=1, Cluster= 3, eps = 1, min sample = 2 |
| Silhouette Score | 0.54 | Silhouette Score | 0.75 |
| David Bouldin Score | 0.72 | David Bouldin Score | 0.95 |
| Hierarchical Clustering | | Density Based Clustering | |
| **b. Sector Classification** | **Cluster =3** | b. Sector Classification | PCA=1, Cluster= 3, eps = 1, min sample = 2 |
| Silhouette Score | 0.54 | Silhouette Score | 0.71 |
| David Bouldin Score | 0.98 | David Bouldin Score | 0.22 |

| Hierarchical Clustering | | Density Based Clustering | |
|---|---|---|---|
| **c. BBLS , PCA Not Applicable** | **Cluster =3** | c. BBLS , PCA Not Applicable | Cluster= 2, eps = .75, min sample = 1 |
| Silhouette Score | 0.38 | Silhouette Score | 0.48 |
| David Bouldin Score | 0.78 | David Bouldin Score | 0.25 |

3. [Financial support for businesses during coronavirus (COVID-19) - GOV.UK (www.gov.uk)](#)
4. Coronavirus Business Impact Loan, 5. Standard Industry Classification

# Appendix 3: Cluster assessment scores for London borough datasets

| Business Demographics Clustering | SIC Based Clustering | BBLS Based Clustering |
|---|---|---|
| PCA=1, Cluster = 2<br>Silhouette score 0.6994104817071428<br>David Bouldin score 0.5791481631597646 | PCA=1 Cluster = 3<br>Silhouette score 0.5469522984952149<br>David Bouldin score 0.4737781120646652 | PCA=1 Cluster = 3<br>Silhouette score 0.5678297507927927<br>David Bouldin score 0.4972965363842175 |
| PCA=1, Cluster = 3<br>Silhouette score 0.5800721244718184<br>David Bouldin score 0.4360738655202177 | PCA=1, Cluster = 2<br>Silhouette score 0.8015192596005515<br>David Bouldin score 0.4593083945397822 | PCA=1 Cluster = 2<br>Silhouette score 0.5629225659568117<br>David Bouldin score 0.5727431414799835 |

**APPENDIX 4 : GEOCODED MAPS OF LONDON BOROUGHS REPRESENTING HIGH, MEDIUM AND LOW IMPACT BOROUGHS FOR BUSINESS BOUNCE BACK LOANS, BUSINESS DEATH RATES AND SECTORAL CONCENTRATION**

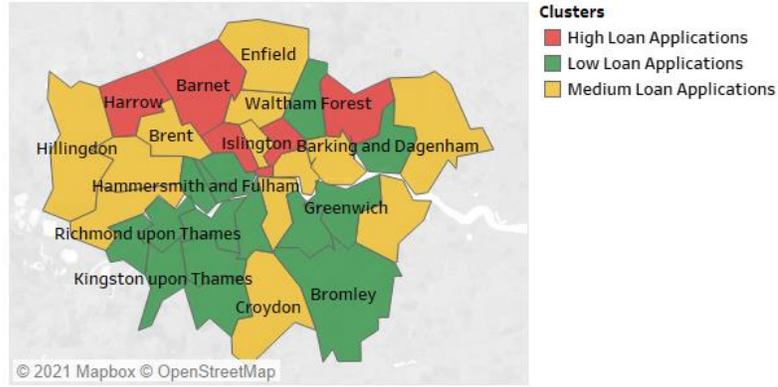

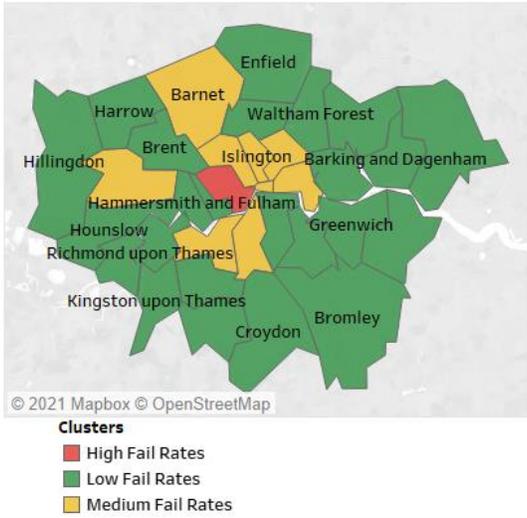

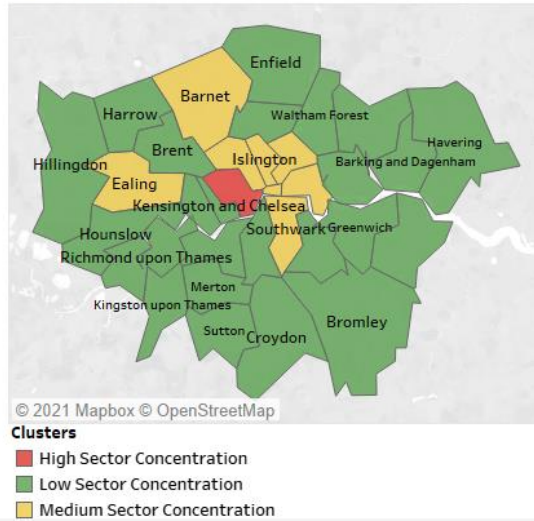

3. Financial support for businesses during coronavirus (COVID-19) - GOV.UK (www.gov.uk)
4. Coronavirus Business Impact Loan, 5. Standard Industry Classification

## APPENDIX 5: NET BUSINESS GROWTH TIMESERIES OF HIGH IMPACT LONDON BOROUGHS

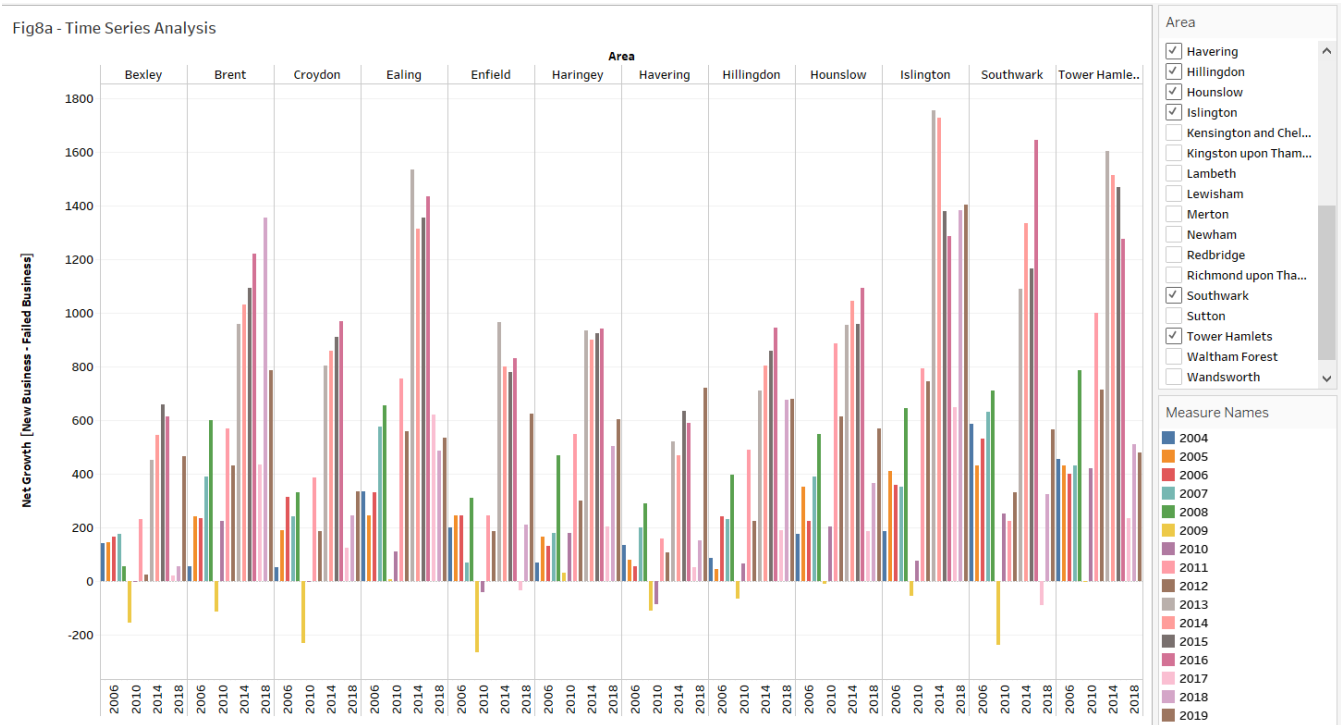

**APPENDIX 6:** LONDON BOROUGH SECTORAL HISTOGRAM PLOTS ENCODED WITH CLUSTER COLORS FOR HIGH, MEDIUM AND LOW IMPACT/ACTIVITY BOROUGHS

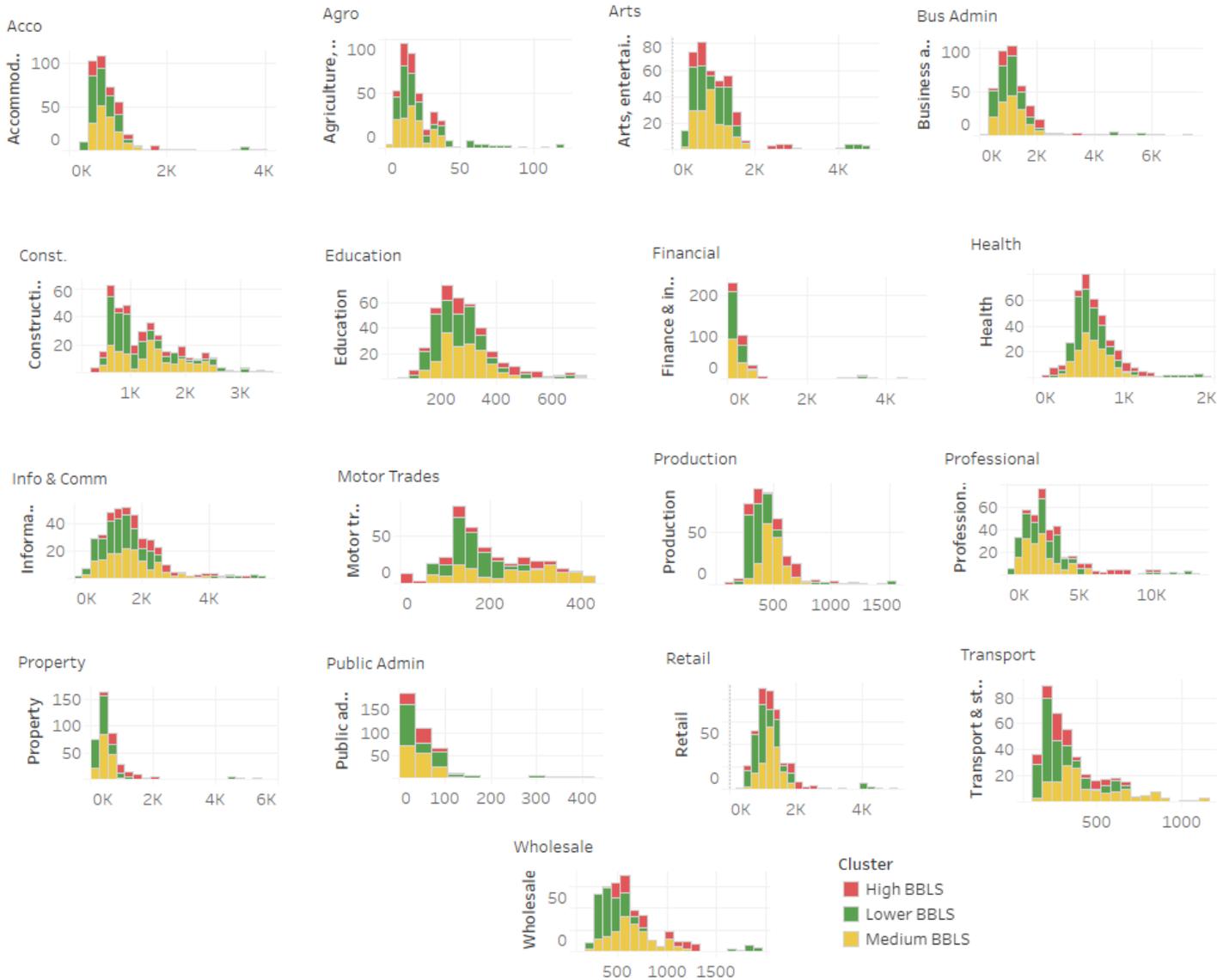

3. [Financial support for businesses during coronavirus (COVID-19) - GOV.UK (www.gov.uk)](#)
4. Coronavirus Business Impact Loan, 5. Standard Industry Classification

# APPENDIX 7 : MANUAL GEOCODE CORRECTIONS
THROUGH MAPPING TABLE

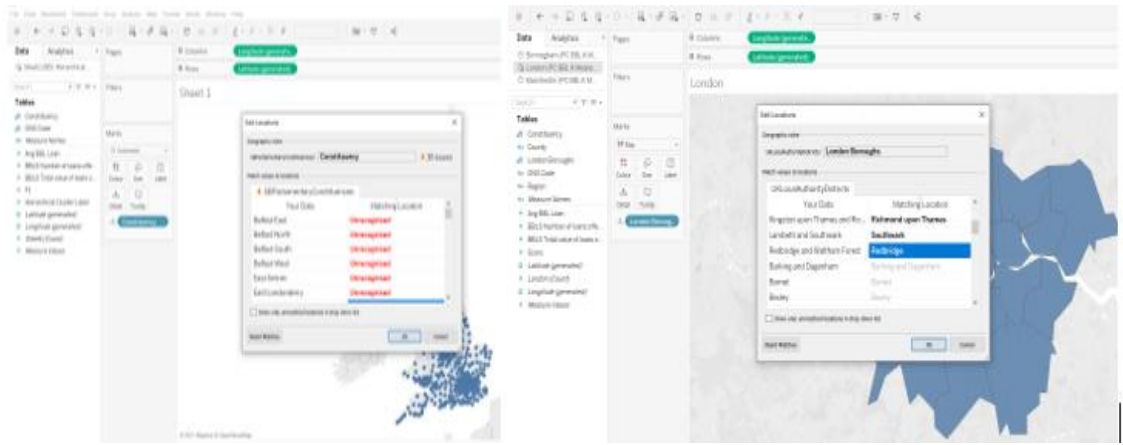

| Original Value | Mapped Value |
|---|---|
| Ards and North Down | North Down and Ards |
| Argyll & Bute | Argyll and Bute |
| Armagh City, Bandridge and | Armagh, Bandridge and Craig…. |
| Bournemouth, Christchurch, Poole | Bournemouth |
| Derry City and Strabane | Derry and Strabane |
| Dorset | North Dorset |
| Dumfried & Galloway | Dumfried and Galloway |
| East Suffolk | Suffolk Coastal |
| Edinburgh City of | City of Edinburgh |
| Herefordshire | Herefordshire, County of |
| Perth & Kinross | Perth and Kinross |
| Scottish Border The | Scottish Borders |
| Somerset County | West Somerset |
| West Suffolk | Mid Suffolk |

| Original Value | Mapped Value |
|---|---|
| Belfast East | 54.5999976 , -5.858829898 |
| Belfast West | 54.595 , -5.964 |
| Belfast North | 54.6630 , 5.9650 |
| Belfast South | 54.5968 , -5.9254 |
| East Antrim | 54.7195, 6.2072 |
| East Londonderry | 54.9460, 6.9530 |
| Fermagh and South Tyrone | 54.5420, 7.3090 |
| Foyle | 55.026 -7.396 |
| Lagan Valley | 54.5359, 5.9772 |
| Mid Ulster | 54.6411, 6.7523 |
| Newry and Armagh | 54.1751, 6.3402 |
| North Antrim | 55.1210, 6.3290 |
| North Down | 54.6536, 5.6725 |
| South Antrim | 54.7210, 6.2410 |
| South Down | 50.916663 -0.499998 |
| Strangford | 54.3698, 5.5557 |
| Upper Bann | 54.4217, 6.3961 |
| West Tyrone | 54.5220, 7.5000 |

APPENDIX 8 : LONDON BOROUGH SECTORAL
TIMESERIES PLOT WITH BUSINESS SENTIMENT
TIMESERIES

Fig8a - Sectoral Concentration 2009-2020

Fig8b - COVID19 Business Sentiments

3. Financial support for businesses during coronavirus (COVID-19) - GOV.UK (www.gov.uk)
4. Coronavirus Business Impact Loan, 5. Standard Industry Classification